\documentclass{llncs}
\usepackage[utf8]{inputenc} 
\usepackage[T1]{fontenc}    
\usepackage{hyperref}       
\usepackage{amsmath}
\usepackage{graphicx}
\usepackage{placeins}
\usepackage{wrapfig}
\usepackage{xcolor}

  \hypersetup{ %
    pdftitle={Interpretable Deep Learning in Drug Discovery},
    pdfauthor={Kristina Preuer et al.},
    pdfsubject={Interpretable Machine Learning, Deep Learning, Neural Networks, Drug Development, Target Prediction},
    pdfkeywords={},
}

\newcommand{\bp}{\mathbf{p}}

\newcommand{\bh}{\mathbf{h}}

\newcommand{\bx}{\mathbf{x}}
\newcommand{\bxp}{\mathbf{x'}}

\newcommand{\bW}{\mathbf{W}}

\newcommand\blfootnote[1]{%
  \begingroup
  \renewcommand\thefootnote{}\footnote{#1}%
  \addtocounter{footnote}{-1}%
  \endgroup
}

\title{Interpretable Deep Learning in Drug Discovery}

\author{
 Kristina Preuer \inst{1}
 \and
 Günter Klambauer \inst{1}
 \and
 Friedrich Rippmann \inst{2}
 \and
 Sepp Hochreiter \inst{1}
 \and
 Thomas Unterthiner \inst{1}
}
\institute{
 LIT AI Lab \& Institute for Machine Learning,\\
 Johannes Kepler University Linz,
 A-4040 Linz, Austria
 \and
Computational Chemistry \& Biology, Merck KGaA, \\ 64293 Darmstadt, Germany
}

\authorrunning{K. Preuer et al.}

\begin{document}
\maketitle

\begin{abstract}
Without any means of interpretation, neural networks
that predict molecular properties and bioactivities are merely black boxes. We will
unravel these black boxes and will demonstrate approaches
to understand the learned representations which are hidden inside these models.
We show how single neurons can be interpreted as classifiers which
determine the presence or absence of pharmacophore- or toxicophore-like structures, thereby
generating new insights and relevant knowledge for chemistry, pharmacology and biochemistry.
We further discuss how these novel pharmacophores/toxicophores can be determined from the network by
identifying the most relevant components of a compound for the prediction of the network.
Additionally, we propose a method which can be used to extract new pharmacophores 
from a model and will show that these extracted structures are consistent with 
literature findings. We envision that having access to such interpretable knowledge 
is a crucial aid in the development and design of new pharmaceutically active molecules, 
and helps to investigate and understand failures and successes of current methods.
\end{abstract}

\section{Introduction}
The central goal of drug discovery research is to identify molecules
\blfootnote{To be published in "Interpretable AI: Interpreting, Explaining and Visualizing Deep Learning"}
that act beneficially on the human (or animal) system, e.g., that have a certain therapeutic effect against particular diseases.
It is generally unknown how chemical structures have to look like in order to induce the wanted biological effects. Therefore,
a large number of molecules have to be investigated to find a potential drug, leading to long drug identification times, and high costs.
This is typically done by means of High-Throughput Screening (HTS), where a biological screening experiment
is used to identify whether a molecule at a given concentration exhibits a certain
biological effect or not. However, running a large number of these experiments is expensive and time-intensive.
Therefore, using computational models as a means of ``Virtual Screening'', i.e.
to predict these biological effects using computational methods and thereby avoiding physical screening, 
has a long tradition in drug development\,\cite{bib:Lionta2014,bib:Lavecchia2015}.

In the past years, the advent of deep learning has allowed neural networks to become the
best-performing method for predicting biological activities based on the chemical structure of the molecules\,
\cite{bib:Ma2015,bib:Mayr2018} mostly because
of their ability to exploit the multi-task setting \cite{unterthiner2014multi}.
Recently, deep learning enabled automated molecule generation \cite{bib:segler2017,bib:olivecrona2017,bib:tsuda2017}, which
has become a new interesting application in the field of drug design.
However, some generative models still have problems with mode collapse \cite{bib:unterthiner2017} and
are hard to evaluate \cite{bib:preuer2018FCD}. Interpretability of neural networks both
for predictions and automated drug design could further push their performance, would
increase their usability and would especially improve acceptance. 

Nowadays, mainly two types of deep neural networks are most frequently used in Virtual Screening: descriptor-based 
feed-forward neural networks (see Section~\ref{sec:FNN}) and graph convolutional neural 
networks (see Section~\ref{sec:GCNN}). Descriptor-based neural networks rely on predefined features, 
so-called molecular descriptors, whereas graph convolutional 
neural networks learn a continuous representation directly from the molecular graph. 
Neural networks take these discrete or numerical representation of a chemical molecule and
calculate their prediction by feeding that representation through several
layers of non-linear, differentiable transformations with many, often millions of, adjustable parameters.
Unfortunately, the function that is encoded in such a neural network is typically impossible to interpret by humans. 
In other words: how the neural network reaches a conclusion is usually beyond the understanding of a human user.
This work aims at bridging this gap in our understanding of neural network predictions for drug discovery. 
Although \cite{bib:hansen2011,bib:Baehrens2010,bib:Schuett2017} have already focused on the difficult question 
how machine learning models can be interpreted, none of these works focus on an in depth analysis of both descriptor 
based and graph based deep learning models for QSAR predictions.

In this work, we first show how a trained neural network can be used to interpret which parts
of a molecule are important for its biological properties, and then demonstrate how 
graph convolutional neural networks can be used to extract annotated chemical substructures, 
known as pharmacophores or toxicophores. 
We will empirically show that neural networks rely on pharmacophore-like features
to reach their conclusions, similar to how a human pharmacologist would. Concretely, we
will show in \autoref{sec:HiddenInterpretation} that the units that form the layers of a neural network 
are pharmacophore detectors. Furthermore, we will demonstrate in \autoref{sec:InputInterpretation} how indicative 
substructures can be determined for individual samples. In the second part of our analysis we will focus on graph 
convolutional neural networks and show that the identified pharmacophores extracted directly from the network match 
well-known, annotated substructures from the literature.

\clearpage
\begin{wrapfigure}{r}{5cm}
	\includegraphics[scale=0.8]{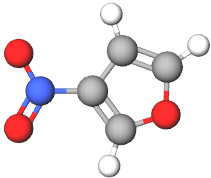}
	\caption{Example of a 2 dimensional structure representation of a molecule.}
	\label{fig:molecule}
\end{wrapfigure}

\section{Learning from molecular structures}
There are multiple scales at which molecules such as the example shown in Fig.~\ref{fig:molecule} can be represented. Molecules 
can be represented by their molecular formula (1D), by their structural formula (2D), by their conformation (3D), by their mutual
orientation and time-dependent dynamics or combinations of all these \cite{Cherkasov2014}. The choice of the right representation 
is task dependent and crucial for the learning algorithm. Most commonly, molecules are described by  so-called Extended Connectivity 
Fingerprints (ECFPs)\cite{bib:Rogers2010}. These 2D-descriptors represent the 2-dimensional structure as a bit vector indicating the
presence and absence of predefined substructures and showed a high predictive performance 
in \cite{bib:Unterthiner2014,bib:Mayr2016,bib:Preuer2017,bib:Lounkine2012}. Therefore, we will use these as descriptors for 
our first experiments described in section \ref{sec:FNN}. \\
However, newer approaches focus on direct end-to-end methods, where the molecular representation is directly learned from 
the molecular graph \cite{bib:Duvenaud2015,bib:Gilmer2017,bib:Kearnes2016}. These graph convolutional methods learn molecular 
representations during the training process and are therefore able to learn wildcards and flexible substructures. We will 
analyse the representations learned by a graph convolutional network in section \ref{sec:GCNN}.

\section{Descriptor-based Feed Forward Neural Networks}
\label{sec:FNN}
A feed forward neural network consists of several layers of computing units.
Each layer $l$ takes as its input the vector of outputs $\bh^{l}$ of the layer below it.
In the first layer, we use the input data $\bh^{0} = \bx$ instead.
The layer $l$ transforms its input according to some parameterized function to produce its own output
\begin{align*}
\bh^{l+1} = f(\bh^{l}\bW^{l})
\end{align*}
where $f$ is an activation function that is applied to
each element, or ``hidden unit'', $h^{l}_i$ of the vector individually.
Each of these elements can be understood
as a feature detector, which detects the presence or absence of some feature in its inputs.
The nature of that feature is defined by the learned parameters $\bW^{l}$,
but is usually very difficult to interpret,
as the features are typically a highly abstract, non-linear function of the input features.
However, we can show that when learned on typical drug development tasks,
these hidden units encode features that are very similar to features used by
pharmaceutical researchers for decades.

\subsection{Interpreting Hidden Neurons}\label{sec:HiddenInterpretation}
A common way to analyze chemical properties of small molecules is by looking at its structures.
Atoms that are close together often form functional groups, which may have specific roles for binding to the respective biological targets.
These functional groups form larger structures which then are responsible for the biological effect, by modulating a biological target (e.g. a protein or DNA). work together to build reactive centers that steer chemical reactions.
Binding to a specific target can only take place when the necessary active centers are present at exactly
the right locations. The exact configuration of active centers is referred to as a pharmacophore\,\cite{bib:Lin2000}.
In other words, a pharmacophore is a molecular substructure, or a set of molecular substructures that is responsible
for a specific interaction between chemical molecules and biological targets.
It is our hypothesis that the hidden units of a neural network learn to detect pharmacophores.
To investigate this, we employed a strategy similar to \cite{bib:Bau2017}, and trained a network that predicts the toxicology of molecules, using the
data set from the Tox21 Data Challenge\,\cite{bib:Huang2014}.
The data set contains around 12\,000 molecules, for each of which twelve
biological effects were measured in wet lab experiments with binary outcome (``toxic'', ``non-toxic'').
These twelve biological effects served pose a multi-task classification problem.
Deep Learning is the best performing method  for this task\,\cite{bib:Mayr2016,bib:Klambauer2017selu},
and we follow the network architecture outlined in \cite{bib:Klambauer2017selu} for our experiments,
using a network with 4 hidden layers of 1024 hidden units with SELU activation function.
To represent our molecules, we use ECFPs\,\cite{bib:Rogers2010} of radius 1, meaning that
the input representation includes presence/absence calls of single atoms and small substructures with at 
most 5 atoms, but gives no concrete information about larger molecular substructures. The network still 
performed relatively well, with an average AUC over the 12 targets of $0.77$, which would still place it
among the top 10 models in the original Tox21 Data Challenge\,\cite{bib:Mayr2016}.

After training the network, we calculate the activation of the hidden units for all molecules in
the training set, and relate them with presence/absence calls of pharmacophores calculated for the
same molecules. For this, we used pharmacophores known to be relevant in
toxicology\,\cite{bib:Sushko2012}, the so-called toxicophores. 
Starting from all the toxicophores in \cite{bib:Sushko2012}, we filter out those that were
were present in less than 20 of our molecules, leaving us with a total of about 650 toxicophores.
For each hidden unit $i$ and each toxicophore $j$, we then performed a Mann–Whitney U-Test to
see if there was a significant difference in the activations between the molecules
where a given toxicophore was present and the ones where it was absent.
We then looked at correlations that were significant at $p\le0.05$
after adjusting for the multiple testing using a very conservative Bonferoni correction.
This leaves us with a total of $\approx$ 290 toxicophores that were significantly
correlated with hidden units of the network.

\begin{figure}[h]
	\includegraphics[width=\textwidth]{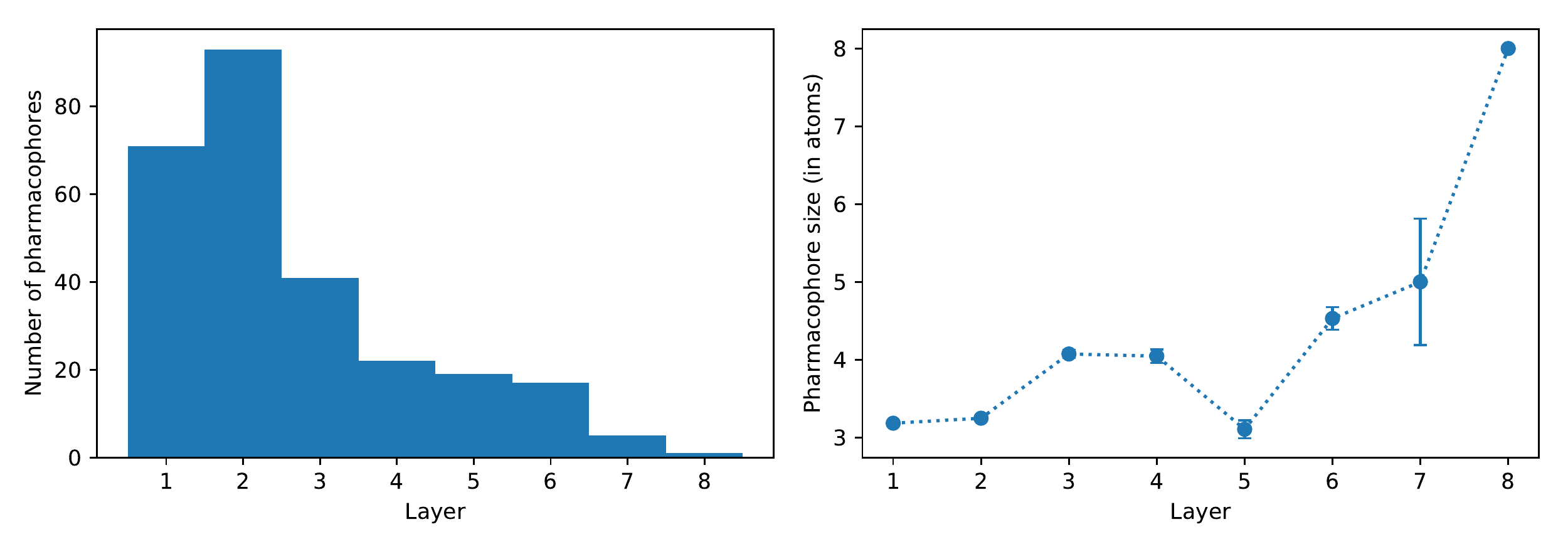}
	\centering
	\caption{\textbf{Left:} How many pharmacophores were found in each layer of the network. \textbf{Right:} Average
    size of the pharmacophore (in number of involved atoms) that are first discovered in a given layer. Error bars
    are standard errors. Note that there are no error bars for layer number 8, as only one pharmacophore was first discovered here.}
	\label{fig:pharmacophores}
\end{figure}

Next, we investigated whether the hierarchy of the layers is associated with the complexity of
the detected toxicophores.
If a network  learns some biologically meaningful information in one of its layers,
it will still need to transport this information through all other layers to use it in its
final prediction at the top layer. This means that every important toxicophore which is discovered
in a lower layer will usually also reappear in all subsequent layers.
Figure~\ref{fig:pharmacophores} shows which layers are discovering our known pharmacophores.
It appears that the pharmacophores are primarily discovered in the first few layers.
The later layers tend to mainly discover pharmacophores that are more complex. Here, we measure
the complexity of a pharmacophore by the number of atoms involved in it. The results are well in line
with the usual view of deep learning constructing more and more complex features in its higher
layers\,\cite{bib:Bengio2013b}.

We have demonstrated that neural networks learn pharmacophore detectors by correlating the hidden units
with known toxicophores. However, not all samples contain a known toxicophore.
Hence, we will demonstrate in the next section how indicative substructures can be identified for any input molecule. 

\subsection{Interpreting the Importance of Input Components}\label{sec:InputInterpretation}
Several ideas have been proposed to explain the predictions from a neural network
by attributing its decisions to specific input features.
See Ancona et~al.\,\cite{bib:Ancona2018} for a short overview of gradient-based methods.
One of these methods is Integrated Gradients\,\cite{bib:Sundararajan2017integratedgradients},
which calculates an attribution value $a_i(\bx)$ for each input dimension $i$.
The value $a_i(\bx)$ can be interpreted as the contribution of $i$ to changing the prediction from
a baseline input $F(\bxp)$ to some specific input $F(\bx)$.
It fulfills many useful properties, for example, it is guaranteed that $F(\bx) - F(\bxp) = \sum_i a_i(\bx)$.
Additionally, Integrated Gradients is the only method that works well even when
there are multiplicative interactions between features of those considered by\,\cite{bib:Ancona2018}. Furthermore, it is independent of the concrete choice of architecture, activation function or other hyperparameters.
The method works by aggregating the gradients of the model's output on the straight
path between $\bxp$ and the target $\bx$. When implementing the method, this integral is approximated
by a sum:
\newcommand{\bxt}{\mathbf{\tilde{x}}}
\begin{align}
a_i(\bx) &= \int_{\alpha=0}^{1} \frac{\partial F\left(\gamma(\alpha)\right)}{\partial \gamma_i(\alpha)} \frac{\partial \gamma_i(\alpha)}{\partial \alpha} \mathrm{d}\alpha \notag \\ 
&= (\bx_i - \bxp_i) \int_{\alpha=0}^{1} \frac{\partial F\left(\bxt \right)}{\partial \bxt_i}\bigg|_{\bxt = \bxp + \alpha(\bx - \bxp)} \mathrm{d}\alpha \notag \\
& \approx (\bx_i - \bxp_i) \sum_{k=1}^{m}\frac{\partial F\left(\bxt \right)}{\partial \bxt_i}\bigg|_{\bxt = \bxp + \frac{k}{m}(\bx - \bxp)}  \frac{1}{m}
\end{align}

where $\gamma(\alpha) = \bxp + \alpha(\bx - \bxp)$ describes the interpolation path and $m$ is the number of steps  in the approximation that controls how exact
the results will be. In our experiments, we obtained good results using an $m$ of $1000$. We used a zero vector as baseline for the feed-forward network, which represents a molecule in which all substructures are set to absent. 

\paragraph{Alcohol Toy Data Set.}
As a proof of concept, we investigate if Integrated Gradients can be
used to extract interpretable pharmacophores from a feedforward neural network.
For this purpose we have constructed a toy data set which classifies
compounds based on a simple rule: compounds containing an alcohol group (i.e. 
a hydroxy group bound to saturated carbon) are classified as positive, whereas compounds
containing no hydroxy groups and carboxylic acids are classified as negative.
The data set consists of 28\,147 samples including 1\,236 positives.
The negative samples comprise 26\,047 oxygen-free molecules and 864 carboxylic acids.
The simplest rule which can be learned is that compounds without hydroxyl
groups are classified as negative. In a second step a rule has to be found
which discriminates between different hydroxyl groups.

In this experiment, we used a fully connected network based on ECFPs
of radius~1. Radius~1 is sufficient for this task, because a
hydroxyl group bound to a saturated carbon can be distinguished from
a carboxylic acid group. In this experiment the model consisted of 4
layers of 1\,024 units with SELU activation and achieved a test set AUC of $>0.99$.
To investigate what was important for the predictions, we used
Integrated Gradients. The feed-forward network is based on ECFPs,
hence Integrated Gradients provide attributions for each fingerprint.
Each fingerprint consists of multiple atoms and one atom is part of
multiple fingerprints. Hence, we calculated the atom-wise
attribution as the sum of the attributions of all fingerprints
in which this atom is part of.

\begin{figure}[h]
	\includegraphics[width=\textwidth]{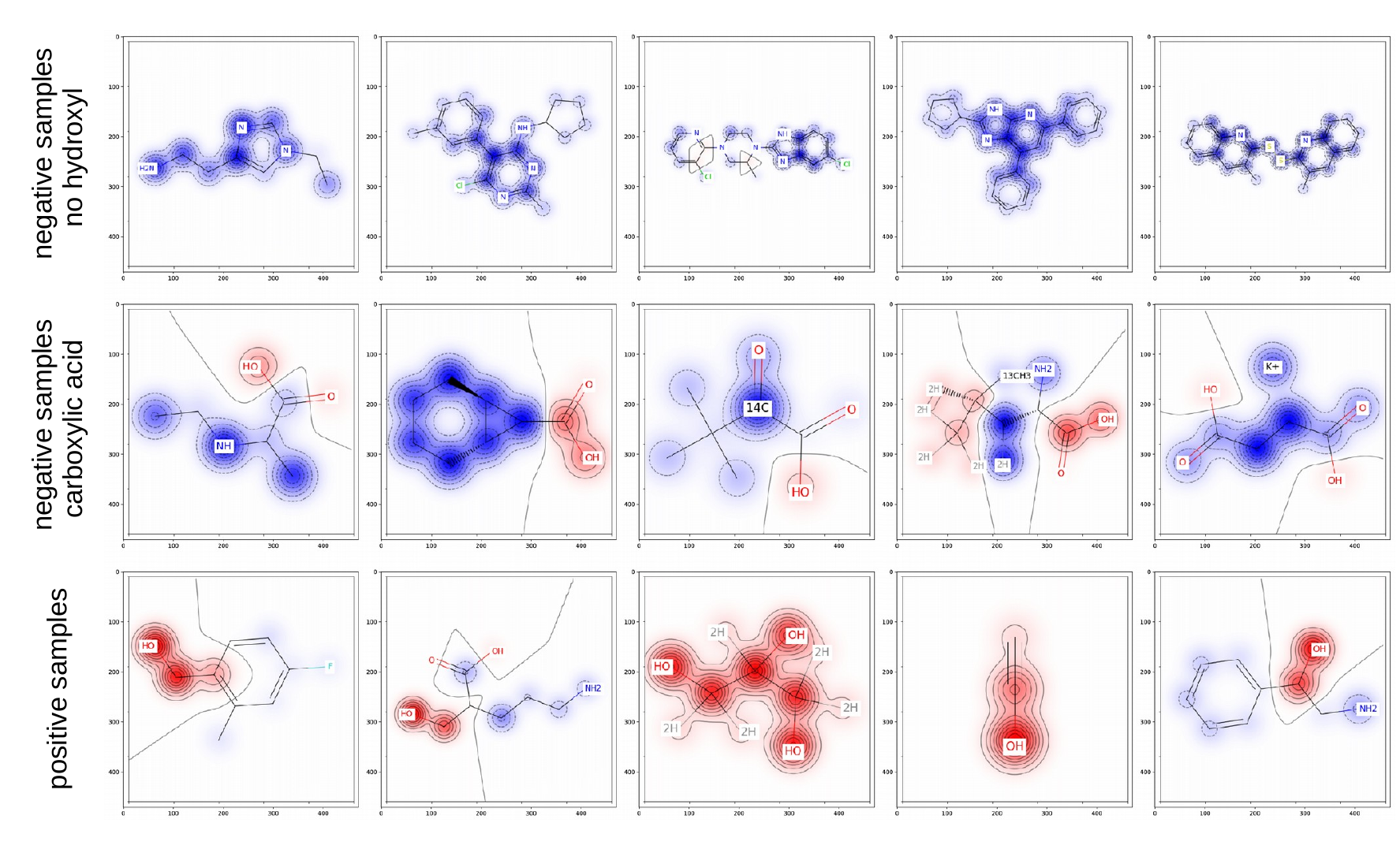}
	\centering
	\caption{Attributions assigned to the atoms by the model for the three
	types of compounds. 5 randomly chosen negative samples without
	hydroxyl groups, negative samples with carboxylic acid groups
	and positive samples are shown in the top, middle and bottom row, respectively.
	Dark red indicates that these atoms are responsible for a positive
	classification, whereas dark blue atoms attribute to a negative classification.}
	\label{fig:alc_ECFP}
\end{figure}

Figure \ref{fig:alc_ECFP} shows the attributions for five randomly
selected molecules for each of the three different molecule types.
In the top, middle and bottom row negative samples without a hydroxyl
group, negative samples with a carboxylic acid group and positive
samples are displayed, respectively. In the first row, almost all atoms
obtain a negative attribution, which is reasonable since non of these
atoms are part of a hydroxyl group. Only in a small fraction (1.9\%) of the 
tested atoms small positive attributions were observed. 
In the second row atoms with carboxylic acids are shown. Atoms not belonging
to the acid group are in general classified as negative, whereas the 
hydroxyl group obtains positive attributions.
This means that the network is still able to identify that the hydroxyl
group is important for a positive classification. In the third row 
molecules with hydroxyl groups are displayed. This group was identified 
as positively contributing to the prediction in 0.83\% of the atoms
by the network and Integrated Gradients. Due to the overlapping fingerprints,
neighboring atoms are also obtaining a slightly positive attribution,
whereas atoms further away are still clearly identified as negative
contributions. This toy example has shown that the Integrated Gradients can
be used to extract the rules underlying the classification.

After this toy experiment, in which we knew which rules had to be applied
to classify the samples, we will investigate whether the decisions for
a more complex task are still interpretable and reasonable.
For this purpose, we used the Tox21 data set.

\paragraph{Tox21 Challenge data set.} In this experiment, we investigated whether Integrated Gradients can be used
to extract chemical substructures which are important for classification
into toxic and non-toxic chemical compounds on the largest available toxicity
data set.

\begin{figure}[h]
	\includegraphics[width=\textwidth]{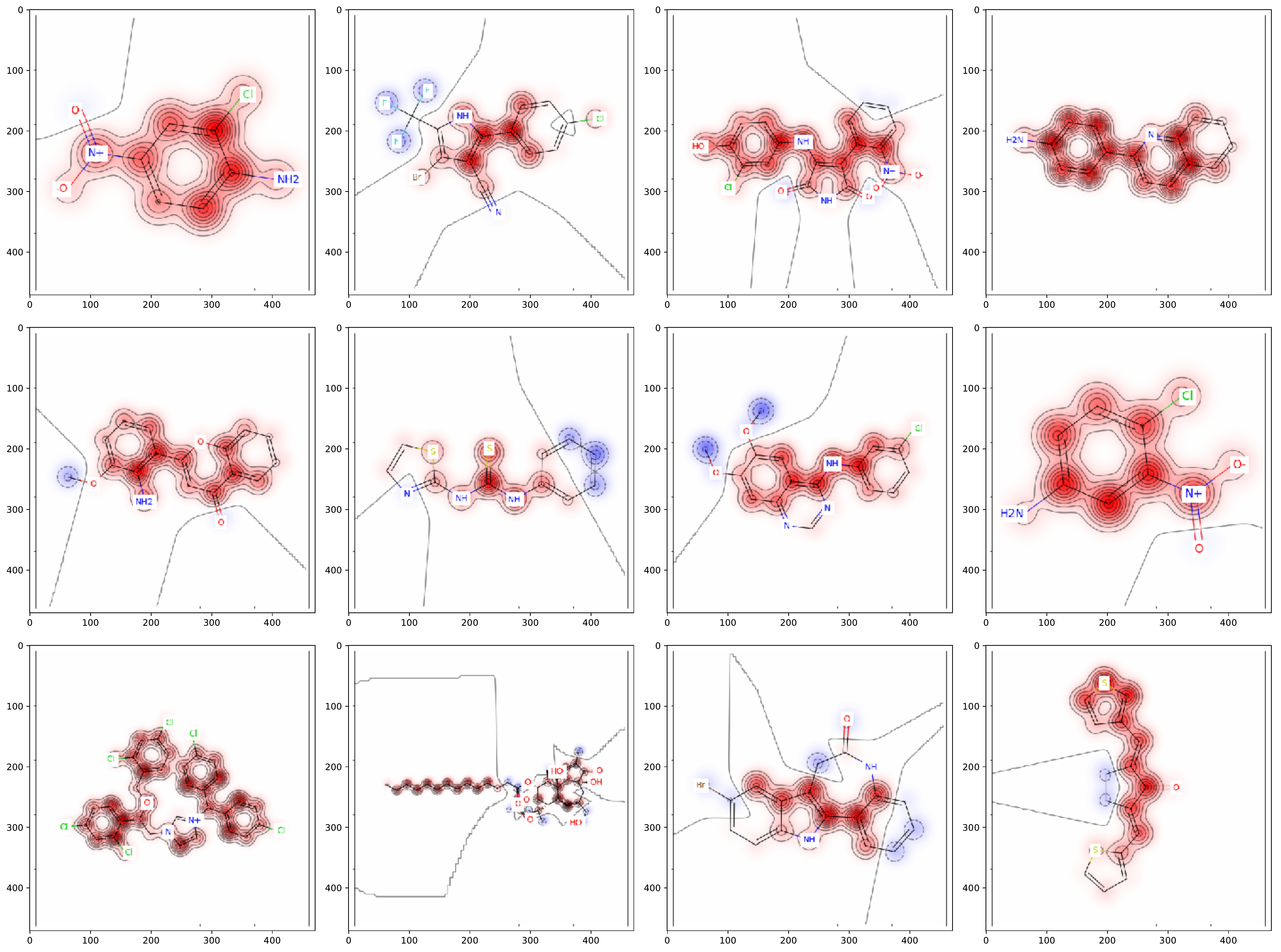}
	\centering
	\caption{Illustration of atom-wise attribution for 12 randomly drawn positive Tox21 samples. Attributions were extracted from a model trained on ECFP\_2 fingerprints. The network clearly bases its decision on larger structures, which were built inside the model out of the small input features.}
	\label{fig:tox_ECFP}
\end{figure}

For this purpose, we trained a fully connected neural network
consisting of 4 SELU layers with 2048 hidden units each 1024 ECFPs
with radius 1 on the Tox21 data set.
This network achieved a mean test AUC of  $0.78$.
We followed the same procedure described above which consists of 
two major steps: applying Integrated Gradients and summarizing feature-wise attributions
into atom-wise attributions. Figure \ref{fig:tox_ECFP} shows 12 randomly drawn,
correctly classified positive test samples. It can be observed that
positive attributions cluster together and form substructures. Please note
that the model was trained only on small substructures, hence the 
formation of the larger pharmacophores is a direct result of the 
learning process. Together with the fact that some attributions
are negative or close to zero, this indicates that the neural
network is able to focus on certain atoms and substructures thereby
differentiating between the indicative and not relevant parts of the input.
The substructures on which the network bases its decision can be viewed as a
pharmacophore-like substructure that indicates toxicity, a so-called ``toxicophore''.

Up to now, we have only considered feed-forward neural networks. In the following sections, we will focus on the 
second prominent networks used for virtual screen: graph convolutional neural networks. We will show how
these networks can be used to extract annotated substructures, such as pharmacophores
and toxicophores, rather than focusing on interpreting individual samples. This knowledge can be 
helpful for understanding the basic mechanisms of biological activities. 

\section{Graph Convolutional Neural Networks}
\label{sec:GCNN}
\begin{figure}[h]
	\includegraphics[width=\textwidth]{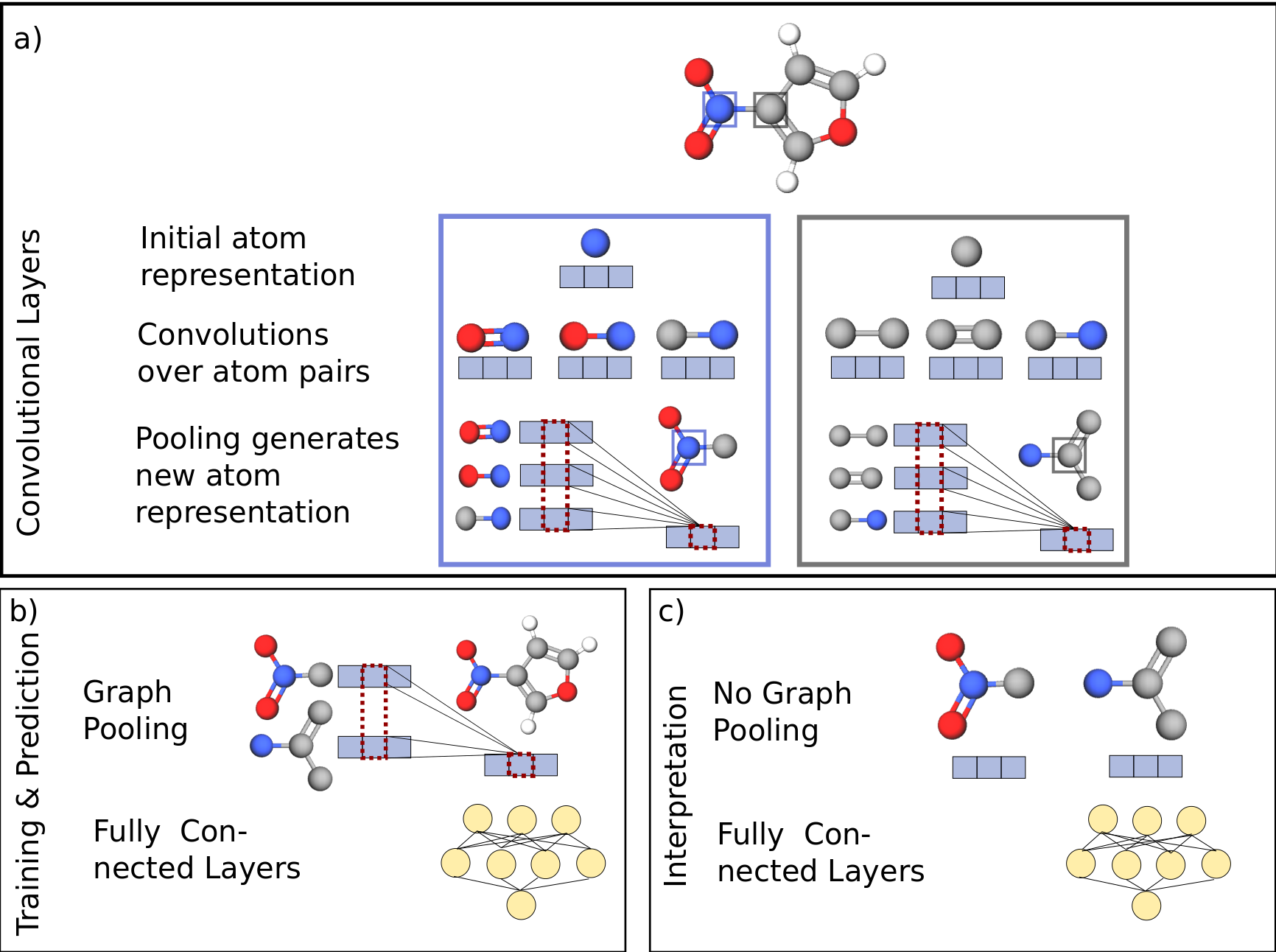}
	\centering
	\caption{a) illustrates the convolutional steps for the blue and the grey center atoms. A new atom representation is formed by convolving over the atom pairs and a subsequent pooling operation. b) shows the graph pooling step which follows the convolutional layers and the fully connected layers at the end of the network. These steps are performed in the training and prediction mode. c) displays the forward pass which is performed to obtain the most relevant substructures. Here, the graph pooling step is omitted and the substructures are directly fed into the fully connected layers.}
	\label{fig:GCNN} 
\end{figure}

We implemented a new graph convolutional approach which is purely based on Keras \cite{bib:Chollet2015} \footnote{The code is available at \url{https://github.com/bioinf-jku/interpretable_ml_drug_discovery}}. 
In our approach, we start similar to other approaches \cite{bib:Duvenaud2015} with an initial atom representation which includes the atom type and the present bond type encoded in a one-hot vector. The network propagates this representation through several graph convolutional layers, whereas each layer performs two steps. In the first step, neighboring atoms are concatenated to form atom pairs. The convolutional filters slide over these atom pairs to obtain a pair representation. The second step is a summarization step, in which a new atom representation is determined. To obtain a new atom representation an order invariant pooling operation is performed over the atom-neighbor pairs. This newly generated representation is then fed into the next convolutional layer. This procedure is illustrated in Fig. \ref{fig:GCNN} a). For training and prediction a summarization step which performs a pooling over the atom representations gives a molecular representation. This step is performed to have a molecular representation with a fixed number of dimensions so that the molecule can be processed by fully connected layers. This steps are shown in Fig. \ref{fig:GCNN} b).\\
Formally, this can be described as following. Let $G$ be a molecular graph with a set of atom nodes $A$. Each atom node $v$ is initially represented by a vector $\bh_v^0$ and has a set of neighboring nodes $N_v$. In every layer $l$ the representations $\bh_v^l$ and $\bh_w^l$ are concatenated $(.,.)$ if $w$ is a neighboring node of $v$ to form the pair $\bp^l_{vw}$. The pair representation $\bh_{p_{vw}}$ is the result of the activation function $f$ of the dot product of the trainable matrix $\bW^l$ of layer $l$ and the pair $p^l_{vw}$. The new atom representation $\bh_v^{l+1}$ is obtained as a result of any order invariant pooling function $g$ such as max, sum or average pooling.
\begin{align}
\bp^l_{vw} &= (\bh_v^l, \bh_w^l)\hspace{1cm}\forall w \in N_v\\
\bh^l_{p_{vw}} &= f(\bp^l_{vw}\bW^l)\\
\bh_v^{l+1} &= g(\{\bh^l_{\bp_{vw}} | w \in N_v\})
\end{align}
A graph representation $\bh^l_G$ of layer $l$ is obtained through an atom-wise pooling step. 
\begin{align}
\label{eq:4}
\bh^l_G = g(\{\bh^l_v | v \in A\})
\end{align}
If skip connections are used, the molecule is represented by a concatenation of all $h^l_G$. 
\begin{align}
\bh_G = (\bh^0_G, \bh^1_G, ..., \bh^L_G)
\end{align}
Otherwise, the representation $\bh^L_G$ of the last layer $L$ is used.
\begin{align}
\bh_G = \bh^L_G
\end{align}
To ensure that our implementation is on the same performance level as other state of the art graph convolutions, we used the ChEMBL benchmark data set. For the comparisons we used the same data splits as the original publication \cite{bib:Mayr2018}, therefore our results are directly comparable. Our approached achieved an AUC of $0.714 \pm 0.15$, which is a 3\% improvement compared to the best performing graph convolutional neural network. 

\subsection{Interpreting convolutional filters}
In this section, we aim at extracting substructures that induce a particular biological effect. This information is useful for designing new drugs and for identifying mechanistic pathways.\\
In graph convolutional neural networks the convolutional layers learn to detect indicative structures, and later fully connected layers combine these substructures to a meaningful prediction. Our aim is to extract the indicative substructures which are learned within the convolutional layers. This can be achieved by skipping the atom-wise pooling step of Eq.~\ref{eq:4} and propagating the atom representation $h_v^l$ through the fully connected layers as depicted in Fig. \ref{fig:GCNN} c). Please note, that we can skip this step, because pooling over the atoms results again in a graph representation which has the same dimensions as the single atom representations. Therefore, we can use the feature representations of individual atoms in the same way as graph representations. Although the predictions are for single atoms, each atom was influenced by its proximity, therefore the scores can be understood as substructure scores. The receptive field of each atom increases with each convolutional layer by one, therefore $h^l_v$ represents substructures of different sizes, depending on how many layers were employed. The substructures are centered at atom $v$ and have a radius equal to the number of convolutional layers. Scores close to one indicate that the corresponding substructure is indicative for a positive label, whereas substructures with a score close to zero are associated with the negative label.

\begin{center}
\begin{figure}
\includegraphics[width=\textwidth]{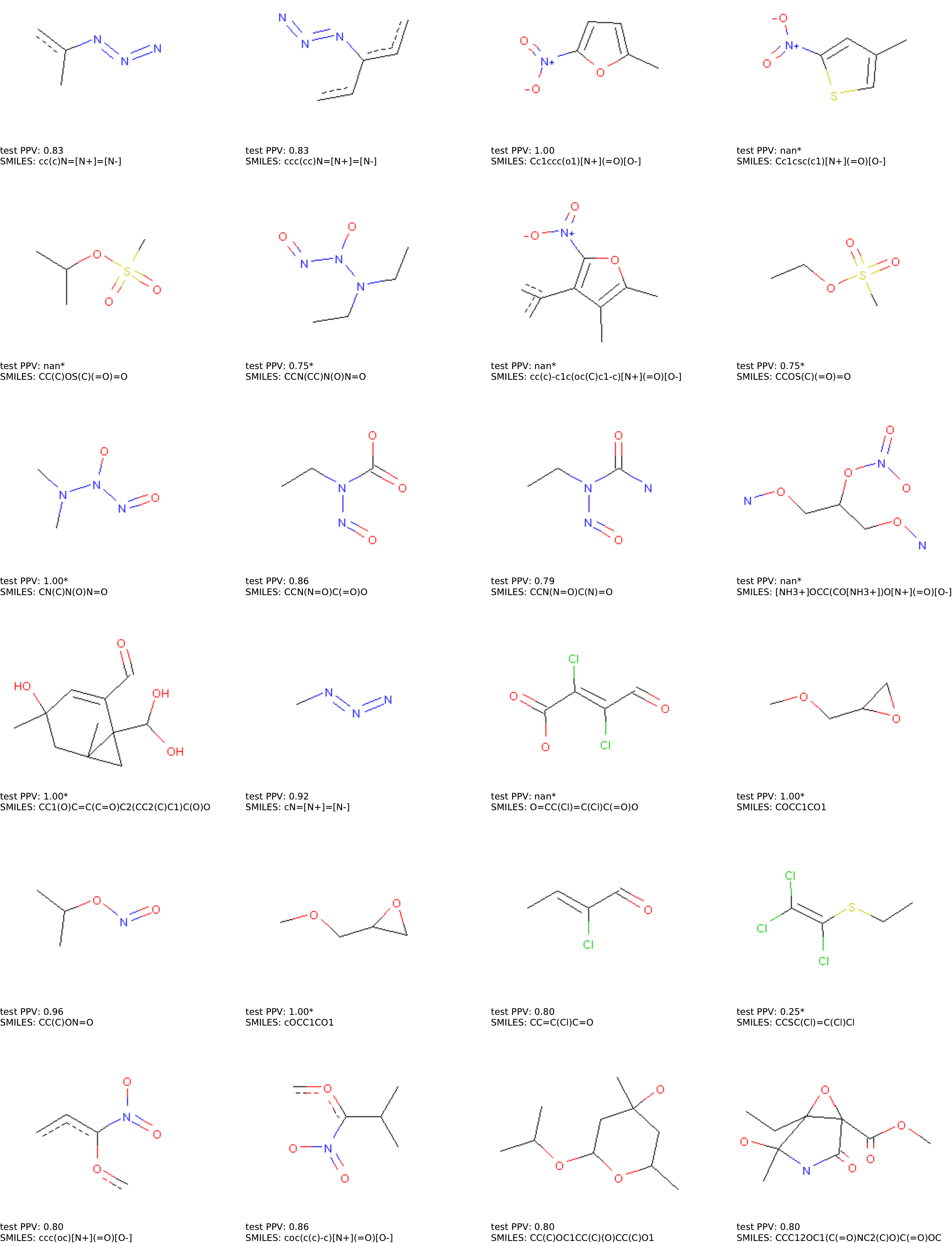}
\caption{This figure displays the structures extracted with our approach from the graph convolutional neural network. Below the structures the corresponding SMILES representation is shown together with the positive predictive value (PPV) on the test set. PPVs which were calculated on less than 5 samples are marked with an asterisk. \label{fig:submols}}
\end{figure}
\end{center}

\begin{center}
\begin{figure}
\includegraphics[width=\textwidth]{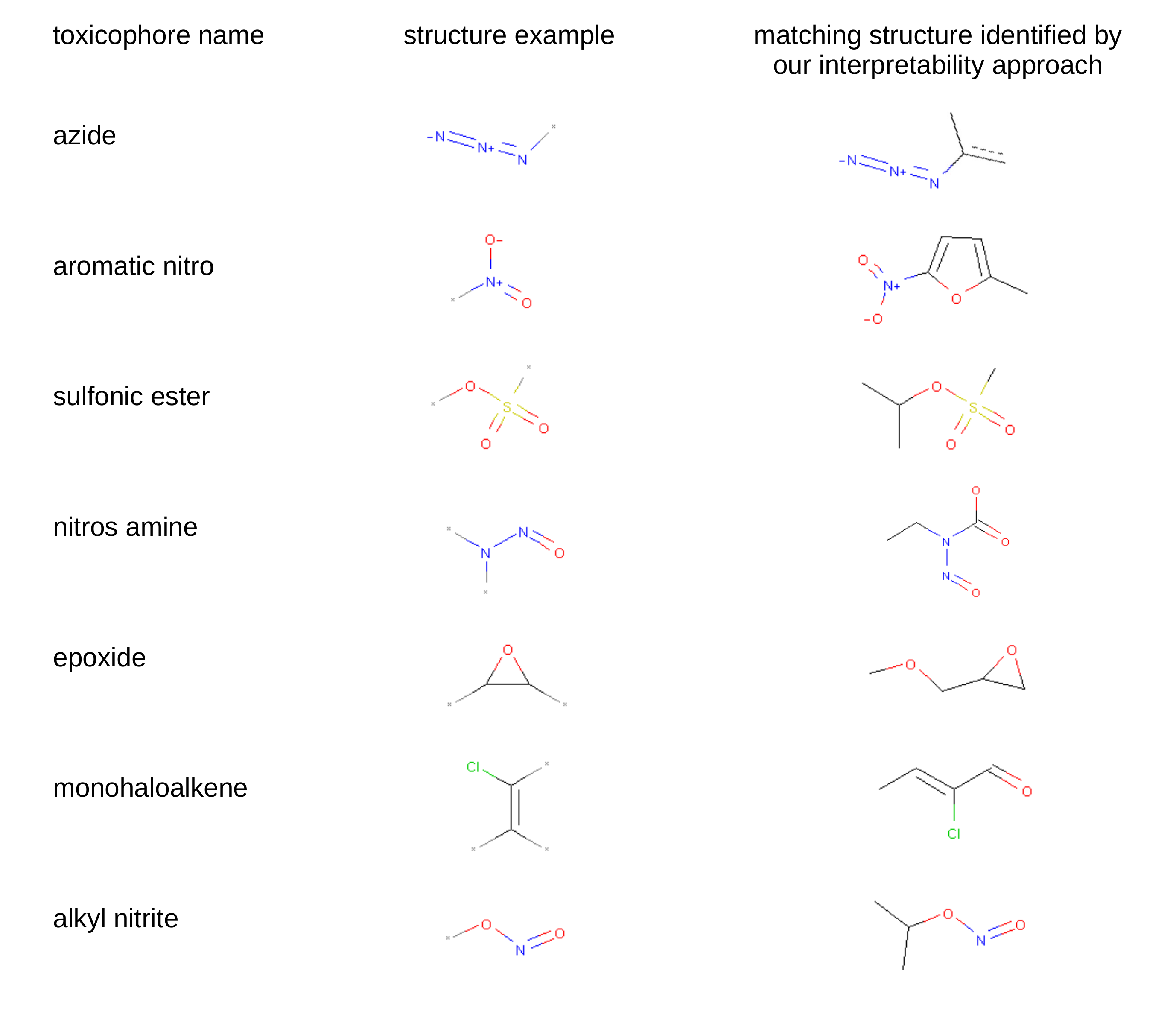} 
\caption{This figure shows annotated SMARTS patterns identified in the literature as mutagenic together with matching structures identified by our interpretability approach. The names of the structures are shown in the first column, the SMARTS patterns found in literature are displayed in the second column and the last column shows a matching example of the top scoring substructures identified by our method.  \label{fig:literature}}
\end{figure}
\end{center}

\paragraph{Ames mutagenicity data set} For this experiment we used a well studied mutagenicity data set. This data set was selected because there exist a number of well-known toxic substructures in the literature which we can leverage to assess our approach. The full data set was published by \cite{bib:Hansen2009}, of which we used the subset originally published by \cite{bib:Kazius2005} as training set and the remaining data as test set. The training set consisted of 4\,337 samples comprising 2\,401 mutagenic and 1\,936 non mutagenic compounds. The test set consisted in total of 3\,315 samples containing 1\,690 mutagens and 1\,625 non mutagens. We trained a network with 3 convolutional layers with 1\,024 filters each followed by a hidden fully connected layer consisting of 512 units. The AUC of the model on the test set was 0.804. \\
To assess which substructures were most important for the network, we propagated a validation set through the network and calculated the scores for each substructure as described above. The most indicative substructures are shown in Figure \ref{fig:submols}. Each substructure is displayed together with its SMILES representation and its positive predictive value (PPV) on the test set. These extracted substructures coincide very well with previous findings in the literature \cite{bib:Kazius2005,bib:Plonik2016,bib:Yang2017}. In Figure \ref{fig:literature} some genotoxic structures found in the literature are displayed with a matching substructure identified by our method. The extracted substructures are known to interact with the DNA. Most of the them form covalent bonds with the DNA via uni-molecular and bi-molecular nucleophilic substitions (SN-1 and SN-2 reactions). Subsequently these modifications can lead severe DNA damage such as base loss, base substitutions, frameshift mutations and insertions \cite{bib:Plonik2016}.\\
Within this section we have demonstrated that our method for interpreting graph convolutional neural networks yield toxicophores consistent with literature findings and the mechanistic understanding of DNA damage. 

\section{Discussion}
Having a black box model with high predictive performance
is often insufficient for drug development: a chemist is not able to derive an actionable hypothesis from just a classification of a molecule into “toxic” or “not toxic”. However, once a chemist “sees” the structural elements in a molecule responsible for a toxic effect, he immediately has ideas how to modify a molecule to get rid of these structural elements and thus the toxic effect. Therefore it is an essential goal is to gain additional knowledge and therefore it
is necessary to shed light onto the decision process within the neural network and to retrieve
the stored information. In this work, we have shown that the layers
of a neural network construct toxicophores and that larger
substructures are constructed in the higher layers. Furthermore,
we have demonstrated that Integrated Gradients is an adequate 
method to determine the indicative substructures in a given molecule. 
Additionally, we propose a method to identify the learned toxicophores 
within a trai,ned network and demonstrated that these extracted 
substructures are consistent with the literature and chemical mechanisms.

\FloatBarrier
\bibliographystyle{splncs04}
\bibliography{bibliography}

\end{document}